\documentclass[10pt]{scrartcl}       %Options: fleqn,draft
\usepackage[bottom=2.4cm,top=2.4cm,left=2cm,right=2cm]{geometry}

\usepackage{amsmath,amssymb,amsthm,graphicx,cases,float,wrapfig,multicol}
\usepackage{abstract,amsfonts,amsbsy,amssymb,amstext,mathrsfs,latexsym,color,url,amsthm,xcolor,authblk,fontenc,bm,mathtools}
\definecolor{NUSBlue}{RGB}{0,61,124} 
\definecolor{NUSOrange}{RGB}{239,124,0}
\usepackage{tikz}
\def\nudge{.5}
\tikzset{axis/.style={ultra thin, Grey, -latex, shorten <=-\nudge cm, shorten >=-2*\nudge cm}}
\tikzset{line/.style={thick}}

\usepackage[font=small,labelfont=bf]{caption}

\DeclareMathOperator{\diag}{diag}
\DeclareMathOperator{\linhull}{span}
\DeclareMathOperator{\spt}{spt}

\newcommand{\bI}{\mathbf{I}}

\newcommand{\bD}{\mathbf{D}}

\newcommand{\bF}{\mathbf{F}}

\newcommand{\bW}{\mathbf{W}}

\newcommand{\ba}{\mathbf{a}}
\newcommand{\bb}{\mathbf{b}}

\newcommand{\bA}{\mathbf{A}}
\newcommand{\bG}{\mathbf{G}}
\newcommand{\bw}{\mathbf{w}}
\newcommand{\bx}{\mathbf{x}}
\newcommand{\by}{\mathbf{y}}

\newcommand{\R}{\mathbb R}

\newcommand{\N}{\mathbb N}

\theoremstyle{plain}
\newtheorem{thm}{Theorem}
\newtheorem{prop}[thm]{Proposition}

\newtheorem{exam}[thm]{Example}
\newtheorem{cor}[thm]{Corollary}

\begin{document}
\renewcommand*{\Authsep}{, }
\renewcommand*{\Authand}{, }
\renewcommand*{\Authands}{, }
\renewcommand*{\Affilfont}{\normalsize\normalfont}   
\setlength{\affilsep}{2em}   % set the space between author and affiliation

\title{Deep Representation with ReLU Neural Networks}

\author[1]{Andreas Heinecke}
\author[2]{Wen-Liang Hwang\thanks{Corresponding author\\
Email addresses: \texttt{whwang@iis.sinica.edu.tw} (Wen-Liang Hwang), \texttt{andreas.heinecke@yale-nus.edu.sg} (Andreas Heinecke)}}
\affil[1]{Yale-NUS College, Singapore 138527, Singapore} 
\affil[2]{Institute of Information Science, Academia Sinica, Taipei 11529, Taiwan}
\date{}
\maketitle

\vspace{-1cm}
\begin{abstract}
We consider deep feedforward neural networks with rectified linear units from a signal processing perspective. In this view, such representations mark the transition from using a single (data-driven) linear representation to utilizing a large collection of affine linear representations tailored to particular regions of the signal space.  
This paper provides a precise description of the individual affine linear representations and corresponding domain regions that the (data-driven) neural network associates to each signal of the input space. In particular, we describe atomic decompositions of the representations and, based on estimating their Lipschitz regularity, suggest some conditions that can stabilize learning independent of the network depth. 
Such an analysis may promote further theoretical insight from both the signal processing and machine learning communities.
\end{abstract}

\section{Introduction}
After having brought about impressive and revolutionary results in machine learning tasks from computer vision, speech recognition or machine translation,
deep neural networks (DNNs) have also been entering into the realms of signal processing.
Deep feedforward neural networks can be viewed as a cascading of affine linear transforms and nonlinear activation functions, producing representations of given data. 
In this view, best visualized via a graph representing the network, the DNN iteratively computes each layer
by transforming the output of the previous layer
with an affine linear operator and a componentwise acting non-linear activation.
From another angle, incepted by the universality theory of shallow neural networks, starting with~\cite{Cybenko1989,Hornik91}, and of deep neural networks, see e.g.\ \cite{Ara18}, DNN's with piecewise linear activation functions can be viewed as piecewise affine linear functions, affine linear on polytopes that partition the input space~\cite{BengioNIPS2014}, that can approximate any function in $L^p(\R^n)$ ($1\leq p\leq \infty$) arbitrarily well. However, the expression power of a DNN cannot be fully leveraged in signal processing without explicit expressions of the affine linear operators, their domains, ranges, and composition from the weight and bias parameters of the network. This paper addresses the expression power of a DNN by providing an explicit formulation of each affine linear mapping and their domains for the case of rectifier activations. In Section~\ref{SPHistory}, we discuss how DNNs with piecewise linear activations may be considered a most significant modern advancement to the long history of signal processing via linear transforms, marking the transition from universal and data-driven linear transforms to data driven piecewise linear transforms. In Section~\ref{Data-driven expression}, we provide a detailed analysis of those piecewise linear transforms for deep feedforward rectifier neural networks.\\
 The main contributions of this paper 
are a configuration expression, that specifies explicitly the hyperplane constraints that bound the domains of each affine linear map and how those refine the input space by increasing the number of layers, as well as an atomic decomposition (Theorem~\ref{Thm3}) for the respective affine maps. 
This characterization of the affine linear pieces unravels precisely how, depending on the region of the input space, the in- and output layers of the network determine the atoms of the representation, and how those atoms are linearly combined over many possible paths through the hidden weights of the network. The precise domain specification and atomic decomposition may facilitate new analytic insight to architectural questions, but also to optimization procedures and empirically successful methods, such as  BatchNorm~\cite{Ioffe2015BatchNA}, dropout~\cite{HintonDropout1} or residual learning \cite{He16}. As an indication we give an estimate of the Lipschitz regularity of the atomic decomposition.  While being important as a characteristic of the representation itself, we also relate it to the smoothness of the gradient of the  networks loss function that governs gradient based training algorithms for DNNs.\\

\section{From orthonormal bases to data-driven representations and deep neural networks}\label{SPHistory}

Many problems of science and engineering can be described by the model
$
\by=\mathcal M(\bx),
$
with input data/signals $\bx\in\R^n$, output data/signals $\by\in\R^m$ and a linear or non-linear operator $\mathcal M$ modelling some process.
Among its many instances, it may for example describe an ill-posed inverse problem where one wishes to reconstruct a certain well structured $\bx$ from an observed $\by$; or a transform, where one wishes to derive some ``good" representation $\by$ of the data $\bx$. The measurement process or transform $\mathcal M$ often contains a linear/non-linear component subject to constraints stemming from, say, physics or engineering. A classic instance is the phaseless reconstruction problem, in which one observes only the modulus of linear Fourier coefficients, thus being an inverse problem consisting of an analysis with a non-linear measurement process. Another instance is synthesis of linear measurements with prior information, e.g.,  sparsity of wavelet frame coefficients in imaging. 
%One interprets $\bx$ as a vector of coefficients that linearly combine frame elements as $\by=T^*\bx$. Among the many solutions that exist due to the non-trivial null-space of $T^*$, one is looking for sparse solutions $\bx$. 
One may also wish to design $\mathcal M$ such that the reconstruction $\by\mapsto \bx$ becomes possible, stable and/or fast. For example, in compressed sensing~\cite{CRT05,Don06} one is interested in designing sensing matrices that allow the recovery of sparse vectors from significantly fewer linear measurements than the signal dimension.

\medskip
\textit{Orthonormal bases and frames:}
For centuries, conventional wisdom suggested that, whenever possible, one should use an orthonormal basis to represent signals. Different orthonormal bases may allow for sparse representations of certain classes of data. The most classic example is the Fourier basis, given by the columns of the matrix
\[
\bF%=[\be_1,\ldots,\be_N]
=N^{-1/2}(\exp(-2\pi i jk/N))_{j,k=0,\ldots,N-1},
\] with the help of which many oscillatory signals become sparsely represented, allowing insight into many phenomena of physics and chemistry. As the Fourier basis is orthonormal, the coefficients of the representation $\by=\bF \bx$, simply given via conjugate transpose, are $\bx=\bF^{*}\by$.\\
In many situations 
orthonormal bases are far from the ideal choice for a representation and it can have great advantages to give up the linear independence imposed on the elements of orthonormal bases. Frames~\cite{DS1952} are advancements of orthonormal bases, derived by relaxing Parseval's identity to a pair of inequalities: A matrix $\bF$ is the synthesis matrix of a \emph{frame}, if there are constants $0<A\leq B$ such that
\begin{align}\label{frame}
A\|\by\|_{2}\leq\|\bF^*\by\|_{2}\leq B\|\by\|_{2}
\text{ for all } \by\in\R^m.
\end{align}
Frames are thus precisely those systems for which signals can be
stably reconstructed from linear measurements.
For any frame there are, in general many, \emph{dual frames} $\bG$, which provide perfect reconstruction of the signal from the linear measurements in the sense that
$
\by=\bF \bG^*\by
$
for all $\by$. 
Dual frames can be derived via different incarnations of a duality principle that hinges on exploiting the adjoint nature of the involved operators.
%In this case the systems $\bT$ and $\bS$ are called \emph{dual frames}. 
In case of \emph{tight frames}, i.e., if \eqref{frame} holds with equality, it is possible to choose $\bG=\bF$, but different dual frames can be chosen to optimally adapt to practical considerations such as, say, minimization of quantization errors.\\
A major advantage of frames is that signals from large classes of data may have common structural features that often translate to the fact that choosing an appropriate frame can force a dimensionality reduction in the sense that the data is sparsely representable via the frame. 
In audio processing, time-varying frequencies are captured sparsely via Gabor tight frames~\cite{Gabor}, comprised of translations and modulations of a window function.
In image processing, wavelet frames~\cite{Dau88,MallatBook} of shifts and dilations of fast oscillating zero-mean functions  can be used to compress and process piecewise smooth images using very few significant coefficients. In both examples, orthonormality is usually given up to gain desired properties, e.g., joint time-frequency localization of the generator in case of Gabor frames, or joint smoothness, symmetry and compact support in the case of the generators of wavelet frames. 
%Again, to achieve simutaneously many desired properties for the wavelets (like smoothness, symmetry and compact support) one gives up orthogonality~\cite{DS}.

\medskip
\textit{Sparse representation and dictionary learning:}
Frames that enable sparse representations of signals yield great advantages, for instance in the interpretation and estimation of the main subcomponents in signals.
While particular frames are predestined for certain signal classes, there remain classes of signals that cannot be sparsely represented with off-the-shelf frames, say, comprised of dilations/modulations and translations of a generator.
The sparse representation problem focuses on the synthesis of signals $\by$ from the span of  some overcomplete dictionary $\bD$, derived from signal domain knowledge, via the sparsest coefficient vector $\bx$,~\cite{Chen01}. Formulated as an optimization problem, the task is to 
\begin{align} \label{sparserep}
\text{minimize } \|\bx \|_0 
\text{ subject to }
\by = \bD \bx,
\end{align}
where $\|\bx\|_0$ returns the number of nonzero entries of $\bx$.
To overcome its NP-hardness, this problem is usually relaxed to a convex optimization problem using the $\ell_1$-norm~\cite{Tib96}. Based on this approach many algorithms have been proposed to iteratively approximate solutions of~\eqref{sparserep}, for an overview see~\cite{TroppW10}.\\
The migration of the sparse representation problem~\eqref{sparserep} to the era of  data-driven methods may be marked with the introduction of K-SVD, \cite{Elad1,Elad2}, where a dictionary and sparse coefficients are being simultaneously learned for a set of observations.

\medskip
\textit{Transition to deep neural networks:} There are many ideas and applications in which neural networks have entered into different aspects of signal processing, see, e.g., \cite{Luc18} for an overview of applications to inverse problems in imaging. 
One example is the question whether approximate solutions to the sparse representation problem \eqref{sparserep} can be derived without using computationally expensive iterative algorithms. To this end, \cite{Mou17} treats the inverse problem \eqref{sparserep} as a regression problem based on a deep neural network that is trained on supervised examples of observations and their sparse representations. After training the network, estimates of sparse representations are calculated by a forward pass of new observations through the network.  To give a second example, the DNN method has also been used in compressed sensing. In \cite{CSGenMod2017}  a $k$-sparse solution  is estimated from noisy measurements $\by=\bA\bx$ obtained via a Gaussian sensing matrix $\bA\in\R^{m\times n}$ by solving the problem
$
\min_{\bx^*} \|\bA \bG(\bx^*) - \by\|_2,
$
where $\bG \colon  \R^k \to \R^n$ is a trained DNN.% and $\bx=\bG(\bx^*)$ is the desired estimate.  \\

\medskip
\textit{Deep representations:}
Approaches as described in the previous paragraph suggest that signal representations should further leverage the data-driven approach in order to obtain representations with better estimation and interpretation properties. Deep neural networks may be considered a next step in the historical development of signal representation described above, in the sense that data is no longer represented  via a single linear representation, like an orthonormal basis, a frame or data driven dictionary, but via an entire collection of affine linear representations. In the case of piecewise linear activations each individual representation of the collection is used for one particular region of a polytope partition of the signal space. In the remainder of this paper we study deep feedforward rectifier neural networks from this angle. Specifically, the architecture we consider is as follows.
For a number $L\in\N$ of layers of widths $N_{0},\ldots,N_{L}\in\N$,  a collection of affine linear
operators $\{M_{\ell}\colon \R^{N_{\ell-1}}\to\R^{N_{\ell}} \}_{\ell=1}^L$ and componentwise acting nonlinear activation functions $\{\rho_{\ell}\}_{\ell=1}^{L-1}$  we consider the map $\mathcal M_L\colon\R^{N_0}\to\R^{N_L}$ defined by 
\begin{align*}
\mathcal M_L(\bx) = M_L \circ\rho_{L-1} \circ M_{L-1}\circ\cdots\circ\rho_1\circ M_1(\bx), 
\end{align*}
to which we will refer to as $L$-layer \emph{deep representation}. 
The affine linear map at the $\ell$-th layer is given by $M_{\ell}(\bx)=\bW_{\ell} \bx +\bb_{\ell }$, with linear part given by a weight matrix $\bW_{\ell}\in\R^{N_{\ell}\times N_{\ell-1}}$, representing edge weights in the graph interpretation of $\mathcal{M}_L$ as a feedforward neural network, and affine shift $\bb_{\ell}$, called bias, representing the offsets of the neurons. We refer to $\mathcal X=\R^{N_0}$ as the  input space. 
%and to $M_1$ and $M_L$ as the input and output layers, respectively. 

\paragraph{Notation:}
We denote matrices bold upper case, vectors bold lower case and scalars in normal font. Moreover, we denote by $x_i$, or $(\bx)_i$, the $i$-th entry of the vector $\bx$,
by $\bw_{k,:i}$ the $i$-th column and by $\bw_{k,j :}$ the $j$-th row of $\bW_k$.
The rank one matrix given by the outer product of the column vector $\bw_{k,:i}$ and the row vector $\bw_{k,j:}$ is denoted by $\bw_{k,:i}\otimes\bw_{k,j:}$.
 We use $|\cdot|$ to denote the cardinality of a set, $\spt \bx$ to denote the support of a vector $\bx$, $\bI$ to denote the identity matrix and $\leq$ to denote the pointwise semi-order on $\R^n$.
Finally, for subsets of $\R^n$ we shorten notation by denoting a set of the form $\{\bx\in\R^n\colon M_{\ell}\bx\geq \mathbf{0}\}$ simply by $\{M_{\ell}\bx\geq \mathbf{0}\}$. Throughout, $M_{\ell}\bx$ will be short for $M_{\ell}(\bx)$.

\section{Data-driven expression for ReLU representations}\label{Data-driven expression}

One of the most effective and widely used non-linear activations is the pointwise acting rectifier $\rho (t)  := \max(0,t)$ for $t\in\R$, \cite{HintonReLU,GlorotBordesBengio2011}. We will refer to $\mathcal{M}_L$ as a \emph{rectified linear unit (ReLU) representation} if all its activations $\rho_k$ are set to be this rectifier.
To begin, consider a $3$-layer representation 
\begin{align}\label{ReLuExample}
\mathcal{M}_3=M_3\rho_2M_2\rho_1 M_1
%=\bW_3(\theta_2(\bW_2(\theta_1(\bW_1\bx+\bb_1))+\bb_2))+\bb_3.
\end{align}
with $\rho_1=\rho_2=\rho$ and denote by
\begin{align} \label{ReLUlayer}
\ba_k =\rho_k \by_k  
\end{align}
the output and input of the $k$-th rectifier, $k=1,2$.
Representing the input in terms of the output we have
\begin{align} \label{nonIO}
y_{k, i}=
\begin{cases}
a_{k,i}  &\text{ if  } a_{k,i} > 0 \\
(-\infty,0]  	&\text{ if }a_{k,i} = 0.
\end{cases}
\end{align}
The non-linearity \eqref{ReLUlayer} can be replaced by 
$
\ba_k=\bD_k \by_k, 
$
using a data-dependent diagonal matrix $\bD_k$ 
whose $i$-th diagonal entry is defined as 
\begin{align}\label{defineD}
d_{k, i}=
\begin{cases}
1 	&\text{ if }a_{k,i} > 0 \\
0    &\text{ if }a_{k,i} = 0 
\end{cases}
=
\begin{cases}
1 	&\text{ if }y_{k,i} > 0 \\
%0/1 & \text{ if } y_{k,i} = 0 \\
0    &\text{ else.}
\end{cases}
\end{align}
The first formulation in \eqref{defineD} captures how $\bD_k$ functions as processing a rectifier backward from $\ba_k$, killing the set-valued entries of $\by_k$ in \eqref{nonIO} and preserving the other entries. The second formulation states how $\bD_k$ functions as processing a rectifier forward from its input $\by_k$, letting the positive entries of $\by_k$ pass, while setting to zeros the negative entries. There is an ambiguity in how to set the diagonal entry if $y_{k,i}=0$ and in our definition in this case the diagonal entry is set to zero.
Note that for a $\{0,1\}$-entry diagonal matrix $\bD_k$, \eqref{defineD} is equivalent to imposing the conditions 
\begin{align} \label{i}
 &\mathbf{0} \leq \bD_k\by_k, \\ \label{ii}
&(\bI-\bD_k)\by_k\leq\mathbf{0} \text{ and } \\ \label{iii}
 &d_{k,i} = 0 \text{ if $y_{k,i} = 0$}. 
 %\text{ and } (\bI-\bD_k)(\b1 + \by_k) > \by_k,
\end{align}
While \eqref{i} excludes the case that $y_{k,i} <  0$ and $d_{k,i} = 1$; \eqref{ii} excludes the case that $y_{k,i} > 0$ and $d_{k,i} = 0$. 
Hence, there does not exist a $\by_k$, such that for any of its components $ 0 <  (\bD_k\by_k)_i \text{ and } ((\bI-\bD_k)\by_k )_i < 0 $. Meanwhile, $y_{k,i} < 0$ and $d_{k,i} = 0$ if and only if $ 0 =  (\bD_k\by_k)_i \text{ and } ((\bI-\bD_k)\by_k )_i < 0 $; as well as $y_{k,i} > 0$ and $d_{k,i} = 1$ if and only if $ 0 <  (\bD_k\by_k)_i \text{ and } ((\bI-\bD_k)\by_k )_i = 0 $.  
%in the intersection that retains $ [\bD_k\by_k]_i = 0$ and $[(\bI-\bD_k)\by_k]_i =  0$ 
We impose  \eqref{iii}, which thus happens if and only if  $(\bD_k\by_k)_i = 0$ and $((\bI-\bD_k)\by_k)_i =  0$. 
%Hence, according to $[\bD_k\by_k]_i$ and $[(\bI-\bD_k)\by_k]_i$, the values of $d_{k,i}$ and $y_{k,i}$ are determined. Meanwhile, $y_{k,i}$ partitions $\R$.
Hereafter, we keep in mind that the diagonal entry corresponding to $y_{k,i} = 0$ is set to zero and neglect  \eqref{iii} to simplify the notation.\footnote{This choice will be rendered irrelevant since it concerns the hyperplane boundary between two regions on which the representation acts affine linear. By continuity of the representation the respective affine linear pieces coincide on those boundaries.}

Working backwards through the non-linearities of the representation, i.e., starting with $\by=M_3\ba_2$ and using $\ba_2 = \rho_2 \by_2 = \bD_2 \by_2$, 
we have $\by= M_3\bD_2\by_2= M_3 \bD_2 M_2\ba_1$, where $\ba_1 = \rho_1 \by_1$. Thus, successively expressing the non-linear relation between out- and inputs of the rectifiers using  data-dependent $\{0,1\}$-entry diagonal matrices, 
the representation \eqref{ReLuExample} becomes
\begin{align*}
\begin{cases}
\by= M_3\bD_2 M_2\bD_1 M_1\bx \\
\text{ with }
d_{1,i}=
\begin{cases}
1 	&\text{ if }y_{1,i} = (M_1\bx)_i > 0 \\
0    &\text{ else}
\end{cases}\\
\text{ and }
d_{2,i}=
\begin{cases}
1 	&\text{ if }y_{2,i} = (M_2\bD_1M_1\bx)_i > 0 \\
0    &\text{ else.}
\end{cases}
\end{cases} 
\end{align*}
or equivalently
\begin{align*}% \label{linearized}
\begin{cases}
\by= M_3\bD_2 M_2\bD_1 M_1\bx  \\
 \text{subject to } \\
\mathbf{0}\le\bD_k\by_k, \, (\bI-\bD_k)\by_k\leq\mathbf{0}; \\
%, \,(\bI-\bD_k)(\b1 +  \by_k) > \by_k; \\
\text{for } \by_k = M_{k}\bD_{k-1} M_{k-1}\cdots M_1\bx \text{ and } k = 1, 2.
 \end{cases}
 \end{align*}

\medskip
The general $L$-layer ReLU representation $\mathcal M_L$ can be expressed as a collection of data-driven affine linear representations:
 \begin{align*}% \label{datadrivenD}
\begin{cases}
\by= M_L\bD_{L-1}\cdots M_2\bD_1 M_1\bx  \\ \text{subject to } \\
\mathbf{0}\le\bD_k\by_k, \, (\bI-\bD_k)\by_k\leq\mathbf{0}; \\
% \,(\bI-\bD_k)(\b1 +  \by_k) > \by_k; \\
\text{for } \by_k = M_{k}\bD_{k-1} M_{k-1}\cdots M_1\bx \text{ and } k = 1, \ldots, L-1.
 \end{cases}
 \end{align*}
We stress that the diagonal matrices are not pre-determined; they are functions of $\by_k$, i.e., depending on the data $\bx$.
The non-linear operator $\mathcal M_L$ is thus expressed as a set of affine linear operators, each of which is determined by the diagonal matrices $\bD_1,\ldots,\bD_{L-1}$, or equivalently by the sign patterns of the input vectors $\by_k$.

 \paragraph{Configuration expression:}
The above description motivates the following terminology and definitions.
%We say that a vector $\theta_k$ is the \emph{diagonal configuration} of a diagonal matrix $\bD_k$ if $\bD_k=\diag(\theta_k)$.
In slight abuse of notation, we call any vector
$\theta=[\theta_1^\top,\ldots,\theta_{L-1}^\top]^\top\in\{0,1\}^{N_1+\cdots+N_{L-1}}$ derived from the concatenation of certain $\theta_k\in\{0,1\}^{N_k}$, $k=1\ldots,L-1$, a \emph{(diagonal) configuration} of the ReLU representation $\mathcal{M}_L$, if the polytope
\begin{align*}%\label{polytope}
R^{\theta}:=
\bigcap_{k=1}^{L-1}
\{ \bx\in\mathcal{X}\colon 
&\mathbf{0}\le\diag{\theta_k}\by_k,  (\bI-\diag{\theta_k})\by_k\leq\mathbf{0},  
%& (\bI-\diag{\theta_k})(\b1 +  \by_k) > \by_k; \nonumber \\
 \text{for } \by_k = M_{k}\diag(\theta_{k-1}) \cdots\diag(\theta_1) M_1\bx
\}
\end{align*} 
is non-empty. 
For a given configuration $\theta$ of $\mathcal{M}_L$, we define the affine linear map 
\begin{align}\label{ReLudexpression}
\mathcal{M}_L^{\theta}
:=M_L\diag(\theta_{L-1})\cdots M_2\diag(\theta_1) M_1
\end{align}
with domain $R^{\theta}$. We will also say that $\mathcal{M}_L^{\theta}$ \emph{induces a configuration}, if $R^{\theta}$ is non-empty.
Then on the restriction to $R^{\theta}$ the ReLU representation $\mathcal{M}_L$  and the affine linear operator $\mathcal{M}_L^{\theta}$ coincide.

\begin{exam}
Let $\theta = [\theta_1^\top, \theta_2^\top]^\top$ be a  configuration of a $3$-layer ReLU representation $\mathcal{M}_3$.
Then the affine linear operator
$
\mathcal{M}_3^{\theta}=M_3\diag(\theta_2) M_2\diag(\theta_1) M_1
$
coincides with $\mathcal{M}_3$
on the convex polytope
%\begin{align}\label{ReLudexpression0}
%\begin{cases}
%\bD^\theta_1M_1\bx\geq 0,          &   (\bI-\bD^\theta_1)M_1\bx\leq 0, \\
%\bD^\theta_2M_2\bD^\theta_1M_1\bx\geq 0,   &   (\bI-\bD^\theta_2)M_2\bD^\theta_1M_1\bx\leq 0.
%\end{cases}
%\end{align}
\begin{align*}%\label{ReLudexpression0}
R^{\theta}=\{\diag(\theta_1)M_1\bx\geq \mathbf{0}\}
\cap  \{(\bI-\diag(\theta_1))M_1\bx\leq \mathbf{0}\} 
% & \cap \{(\bI-\diag(\theta_1))(\b1 +  M_1\bx) > M_1\bx\}  \\
\cap \{\diag(\theta_2)\mathcal{M}_2^{\theta_1}\bx\geq \mathbf{0}\} 
  \cap       \{(\bI-\diag(\theta_2))\mathcal{M}_2^{\theta_1}\bx\leq \mathbf{0}\},
% \\
% &  \cap      \{(\bI-\diag(\theta_2))(\b1+\mathcal{M}_2^{\theta_1}\bx) > \mathcal{M}_2^{\theta_1}\bx\},
\end{align*}
i.e., on the set of all $\bx\in\mathcal X$ that satisfy
\begin{align*}
(M_1 \bx)_i &\in
\begin{cases}
(0,\infty) & \text{ if } i \in \spt\theta_1,\\
(-\infty,0] & \text{ else,}
\end{cases}
\quad\text{ and }\quad
(M_2 \diag(\theta_1)M_1\bx)_i \in
\begin{cases}
(0,\infty) & \text{ if } i \in \spt\theta_2,\\
(-\infty,0] & \text{ else.}
\end{cases}
\end{align*}
If $\mathcal X=\R^2$ and if $\theta = [ 0, 1, 1, 0]^\top\in\R^{2+2}$ is a configuration, then  $\theta$ defines 
\[%M_3\bD_2^\theta M_2 \bD_1^\theta M_1=
\mathcal{M}_3^{\theta}=M_3\diag(1,0)M_2\diag(0,1)M_1\] 
%Then $\bD_1^\theta =\diag(0,1)$ and 
%$\bD_2^\theta = \diag(1,0)$ 
%Thus, $\bD_1^\theta = \begin{pmatrix} 
%0 & 0\\ 0 & 1 
%\end{pmatrix}$, $\bD_2^\theta = \begin{pmatrix} 
%1 & 0\\ 0 & 0
%\end{pmatrix}$. 
on the polytope 
\begin{align*}
\{(M_1\bx&)_1 \leq 0 \}
\cap\{(M_1\bx)_2 > 0\}  \cap 
\{(M_2\diag(0,1)M_1\bx)_1 >  0\}   \cap\{(M_2\diag(0,1)M_1\bx)_2 \leq 0\}.
\end{align*}
\end{exam}

In the remainder of this section we recall how the configurations of a ReLU representation partition the input space in increasingly finer polytopes, before describing in detail the affine linear maps.

\subsection{Input space partition}
Given a ReLU representation $\mathcal{M}_L$, denote by $\Theta_{k}$, for $k=1,\ldots,L-1$, the set of all configurations of $\mathcal M_{k+1}$. Then every configuration in $\Theta_k$ is derived from a configuration in $\Theta_{k-1}$ via concatenation with a vector from $\{0,1\}^{N_k}$. Note however that not 
 all $2^{N_k}$ possible vectors $\theta_k\in\{0,1\}^{N_k}$ are part of a  configuration $[\theta_1^\top,\ldots,\theta_{k}^\top]^\top\in\Theta_{k}$. Lower estimates of the size of $\Theta_k$ are given in~\cite{BengioNIPS2014}.
Whether or not a certain $[\theta_1^\top,\ldots,\theta_{k}^\top]^\top$ is a configuration depends on  $\mathcal{M}_{k}^{[\theta_1^\top,\ldots,\theta_{k-1}^\top]^\top}$. If, say, $M_1 =0$ for some ReLU representation, then $\theta_k$ must be the zero vector for all $k=1,\ldots,L-1$ and hence for such a deep representation there is only one configuration $\theta = [0,\ldots,0]^\top$ possible. We consider a slightly less trivial example in more detail.

\begin{wrapfigure}{r}{0.47\textwidth}
\captionsetup{format=plain}
\vspace{-0.5cm}
\centering
 \resizebox {0.47\textwidth} {!} {
\begin{tikzpicture}
\fill[NUSOrange,opacity=1] (-1.5,-1.5)--(-1.5,-1)--(2.136,0.454)--(5,-1.5);
\fill[NUSOrange,opacity=.7] (3,0.8)--(6,2)--(6,-1.3); 
\fill[NUSOrange,opacity=.85] (2.136,0.454)--(3,0.8)--(6,-1.3)--(6,-1.5)--(5,-1.5);
\fill[NUSOrange,opacity=.45] (2.136,0.454)--(6,2)--(6,4)--(5,4)--(1.6,0.83);
\fill[NUSOrange,opacity=.35] (1.6,0.83)--(5,4)--(-1.5,4)--(-1.5,3);
\fill[NUSOrange,opacity=.2] (2.136,0.454)--(0,1.95) -- (-1.5,1.35) -- (-1.5,-1);
\fill[NUSOrange,opacity=.1] (-1.5,3) -- (0,1.95) -- (-1.5,1.35);
%
%% axis
%\draw[->,thin] (-1.5,0) -- (6,0);
%\draw[->,thin] (0,-1.5) -- (0,4);
%%
\draw[line] (-1.5,-1)--(6,2)
node [black,pos=0.8, below,rotate=23] {$\scriptstyle  \bw_{1,1:},\bx+(\bb_1)_1=0$};
\draw[line] (-1.5,3)--(5,-1.5)node [black,pos=0.75, below,rotate=-35] {$\scriptstyle  \bw_{1,2:}\bx+(\bb_1)_2=0$};
%second layer
\draw[dashed,line] (-1.5,1.35) -- (0,1.95);
\draw[dashed,line] (3,0.8) -- (6,-1.3);
\draw[dashed,line] (1.6,0.83) -- (5,4)
node [black,pos=0.6, above,rotate=43] {$\scriptstyle M_2\diag(\theta_1^3)M_1\bx=0$};
%%\node[rotate=-90] at (2.5,1) {a long text};
%%
\node[above right] at (1,-1.5) {$R^{\theta_1^0}$};
\node[below left] at (6,1) {$R^{\theta_1^2}$};
\node[below right] at (1,4) {$R^{\theta_1^3}$};
\node[below right] at (-1.5,1) {$R^{\theta_1^1}$};
\end{tikzpicture}
}
\caption{Partitioning $\mathcal X=\R^2$ for the $3$-layer ReLU representation $\mathcal{M}_3$ of Example~\ref{ReLUEx2}. The first layer partitions $\mathcal X$ into four unbounded polygons $R^{\theta_1^0}, R^{\theta_1^1}, R^{\theta_1^2}, R^{\theta_1^3}$, divided by the solid lines $(M_1\bx)_1=0$ and $(M_1\bx)_2=0$. Since the image of $M_2$ is one-dimensional, the second layer potentially can further partition each polygon of the first layer into maximally two polygons. The dashed line in, say,  $R^{\theta_1^1}$ is orthogonal to $\bW_2\diag(1,0)\bW_1=w_{2,11}\bw_{1,1:}$, thus parallel to the solid line $(M_1\bx)_1=0$ and shifted due to the bias terms.
The polygon corresponding to the zero configuration is never further partitioned.}
\label{Fig1}
\vspace{-10pt}
\end{wrapfigure}
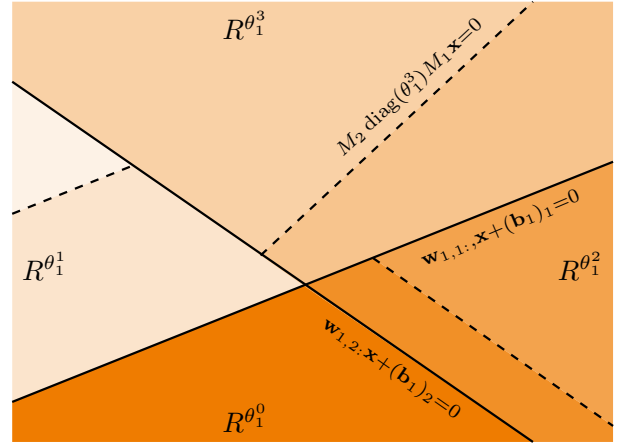

\begin{exam}\label{ReLUEx2}
Consider a ReLU representation $\mathcal{M}_3$ on $\mathcal X=\R^2$ where $M_1\colon\R^2\to\R^2$ is surjective and $M_2\colon \R^2\to\R$. % and suppose $\bW_1$ has full rank. 
%This implies $\bW_1 \in \R^{2 \times 2}$, $\bW_2 \in \R^{2 \times 1}$, $
Then
\[
\Theta_1 =\left\{
\theta_1^0 = \begin{pmatrix} 0\\0 \end{pmatrix},
\theta_1^1 = \begin{pmatrix} 1\\0 \end{pmatrix},
\theta_1^2 = \begin{pmatrix} 0\\1 \end{pmatrix},
\theta_1^3 = \begin{pmatrix} 1 \\ 1 \end{pmatrix}
\right\},
\]
and the input space $\mathcal X$ is first partitioned by the configurations from $\Theta_1$ into the polygons
\begin{align*} 
R^{\theta_1^0} &=\{M_1\bx \leq \mathbf{0}\}, &
R^{\theta_1^1} &=\{(M_1\bx)_1 > 0\}\cap\{(M_1\bx)_2 \leq  0\}, \\
R^{\theta_1^2} &= \{(M_1\bx)_1 \leq 0\}\cap\{(M_1\bx)_2 >  0\}, &
R^{\theta_1^3} &= \{M_1\bx > \mathbf{0}\}.
\end{align*} 
These polygons are further partitioned by the second layer. 
Since $M_2 \diag(\theta_1^0) M_1 = 0$,
the only diagonal configuration that can be achieved via a concatenation from $\{0,1\}$ to $\theta_1^0$ is  $\theta_2^0=[0,0,0]^\top$ and thus $R^{\theta_1^0}$ is not further partitioned. 
The partitions of $R^{\theta_1^j}$, for $j=1,2,3$, are derived depending on the affine transforms $M_2 \diag(\theta_1^j) M_1$.
The polygon $R^{\theta_1^j}$ is  partitioned into the union of the two polygons
$
\{ \bx\in R^{\theta_1^j}\colon M_2 \diag(\theta_1^j) M_1 \bx >  0\} 
$
corresponding to  $[(\theta_1^j)^\top, 1]^\top$ and
$
\{\bx\in R^{\theta_1^j}\colon M_2 \diag(\theta_1^j) M_1 \bx 
\leq 0\}
$
corresponding to $[(\theta_1^j)^\top, 0]^\top$, unless one of those sets is empty, in which case the corresponding vector is not a configuration.  
 Altogether, $\mathcal{X}$ is partitioned into potentially  up to  $7$ convex regions, as illustrated in Figure~\ref{Fig1},
%, coded $R^0, R^2, R^6, R^1, R^5, R^3, R^7$,
 corresponding to the configurations
\begin{align*}
\Theta_2 =
%= \begin{pmatrix} \text{diag }\bD^1 \\ \text{diag }\bD^2 \end{pmatrix}
%= \begin{pmatrix} \Theta_1 \\  \Theta_2 \end{pmatrix}
\left\{
\begin{pmatrix} 0\\0\\0 \end{pmatrix},
\begin{pmatrix} 0\\1\\0 \end{pmatrix},
\begin{pmatrix} 0\\1\\1 \end{pmatrix},
\begin{pmatrix} 1\\0\\0 \end{pmatrix},
\begin{pmatrix} 1\\0\\1 \end{pmatrix},
\begin{pmatrix} 1\\1\\0 \end{pmatrix},
\begin{pmatrix} 1\\1\\1 \end{pmatrix}
\right\},
\end{align*}
but, depending on the actual parameters, a smaller $\Theta_2$ is possible.
Each configuration is associated to an affine linear map via \eqref{ReLudexpression}, to which $\mathcal{M}_3$ is equal to when restricted  to the corresponding polytope. The non-linear operator $\mathcal{M}_3$ is piecewise affine linear, comprised of the (up to)  $7$ affine linear maps
$\mathcal{M}_3^{\theta}$, $\theta\in\Theta_2$. 
%All affine linear mappings use the same coordinate system.
Note that if the bias vectors $\bb_1$ and $\bb_2$ are zero, then $M_1=\bW_1$ and $M_2=\bW_2$, and thus the regions are convex cones arising from halfspace intersections through the origin.
\end{exam}

We record in the following result how the consecutive layers of an $L$-layer ReLU representation $\mathcal{M}_L$ define increasingly finer partitions of the input space.\footnote{To be precise (see comment on \eqref{iii}), here partition has to be understood in the sense that the interiors of the participating sets have empty intersection.} Restricted to each polytope of the final partition, $\mathcal{M}_L$ is equal to an affine linear operator specified by the diagonal configuration corresponding to that region.

\begin{prop}\label{lemReLU}
Let $\mathcal{M}_L$ be a ReLU representation, $k\in\{1,\ldots,L-1\}$ and $\Theta_k$ the set of configurations of $\mathcal{M}_k$. 
\begin{itemize}
\item[(i)] If $\theta=[\theta_1^\top,\ldots,\theta_{k}^\top]^\top\in\Theta_{k}$, then on $R^\theta$ the representation $\mathcal {M}_{k+1}$ coincides with
\[
\mathcal{M}_{k+1}^{\theta}=M_{k+1}\diag(\theta_{k})\cdots M_2\diag(\theta_1) M_1.
\]
\item[(ii)] Define $R^{\Theta_k}=\{R^{\theta}\colon \theta\in\Theta_k\}$. Then
$R^{\Theta_k}$ is a partition of $\mathcal X$ and 
$R^{\Theta_{k+1}}$ is a refinement of $R^{\Theta_{k}}$.
\end{itemize}
\end{prop}
\proof
The first claim follows by induction on the layers from the construction; and so does the second. Indeed, $R^{\Theta_1}$ is a partition of $\mathcal X$. Suppose $R^{\Theta_k}$ partitions $\mathcal{X}$, 
let $\theta \in \Theta_k$ and denote $R^{\theta}$ the domain of the affine map $\mathcal{M}_{k+1}^{\theta}$. %Every vector in $R^{\theta}$ belongs to  exactly one configuration induced by $\mathcal{M}_{k+1}^{\theta}$.
%Hence, 
The entirety of the regions of  the configurations induced by $\mathcal{M}_{k+1}^{\theta}$ partition $R^{\theta}$. Since $\Theta_{k+1}$ is defined as the union of the configurations induced by all $\mathcal{M}_{k+1}^{\theta}$ with  $\theta \in \Theta_k$, the collection $R^{\Theta_{k+1}}$ is a refinement of $R^{\Theta_k}$. 
\qed

\begin{figure}[t] % h for here, t for top, b for bottom, !htbp
\captionsetup{format=plain}
\centering
 \resizebox {\textwidth} {!} {
\begin{tikzpicture}
%\draw [->] (-1.5,0)--(6,0);
%\draw [->] (0,-1.5)--(0,4);
%first layer
\fill[NUSOrange,opacity=1] (-1.5,-1.5)--(-1.5,1)--(1.7937,1.43916)--(5,-1.5);
%%right
\fill[NUSOrange,opacity=0.25] (2.5,1.533)--(3.5,1.667)--(6, -0.613)--(6, -1.5)--(5.81,-1.5);
\fill[NUSOrange,opacity=0.4] (2.5,1.533) -- (5.81,-1.5) -- (5,-1.5)--(1.7937,1.43916); 
\fill[NUSOrange,opacity=0.15] (3.5,1.667) -- (6,2) -- (6, -0.613);
%%left
\fill[NUSOrange,opacity=0.8]  (-1.5,2.883) --(-1.5,4) -- (-1,4)--(0,3.0833);
\fill[NUSOrange,opacity=0.6] (-1.5,2.883)--(-1.5,2.209)--(0.6421,2.495)--(0,3.0833);
\fill[NUSOrange,opacity=0.4](-1.5,2.209)--(-1.5,1)--(1.7937,1.43916)--(0.6421,2.495);
%%upper
\fill[NUSOrange,opacity=0.3]  (-1,4)--(1,2.167)--(5.5,4);
\fill[NUSOrange,opacity=0.4]  (2,4)--(6,4)--(6,2)--(3,1.6);
\fill[NUSOrange,opacity=0.15] (6,4)-- (6,2)--(1.7937,1.43916)--(1,2.167)--(5.5,4);
\draw[line] (-1.5,1)--(6,2);
\draw[line] (-1,4)--(5,-1.5);
%% parallel lines:
\draw[dashed,line] (-1.5,2.883)--(0,3.0833);
\draw[dashed,line] (-1.5,2.209)--(0.6421,2.495);
\draw[dashed,line] (1,2.167)--(5.5,4);
\draw[dashed,line] (3,1.6)--(2,4);
\draw[dashed,line] (2.5,1.533)--(5.81,-1.5);
\draw[dashed,line] (3.5,1.667)--(6,-0.613);
\node[above right] at (1,-1.5) {$R^{\theta_1^0}$};
\node[below left] at (6,0.7) {$R^{\theta_1^2}$};
\node[below right] at (3,4) {$R^{\theta_1^3}$};
\node[below right] at (-1.5,3) {$R^{\theta_1^1}$};
%second layer
\end{tikzpicture}
\hspace{0.2cm}
\begin{tikzpicture}
\fill[NUSOrange,opacity=1]  (0,0)--(-3.75,-0.375)--(-3.75,-2.75)--(-1.833,-2.75);
\fill[NUSOrange,opacity=0.8]  (0,0)--(3.75,0.375)--(3.75,-2.75)--(-1.833,-2.75);
\fill[NUSOrange,opacity=0.6]  (0,0)--(3.75,0.375)--(3.75,1.7);
\fill[NUSOrange,opacity=0.5]  (0,0)--(3.75,1.7)--(3.75,2.2);
\fill[NUSOrange,opacity=0.4]  (0,0)--(3.75,2.2)--(3.75,2.75)--(1.833,2.75);
\fill[NUSOrange,opacity=0.2]  (0,0)--(1.833,2.75)--(-3.75,2.75)--(-3.75,-0.375);
%\draw [->,thin] (-3.75,0)--(3.75,0);
%\draw [->,thin] (0,-2.75)--(0,2.75);
\draw[line] (-1.833,-2.75)--(1.833,2.75)
node [black,pos=0.2, below,rotate=56.5] {$\scriptstyle  \bw_{2,1:}\bx=0$};
\draw[line] (-3.75,-0.375)--(3.75,0.375)
node [black,pos=0.8, below,rotate=5.5] {$\scriptstyle \bw_{1,1:}\bx=0$};
\draw[dashed,line] (0,0)--(3.75,1.7);
\draw[dashed,line] (0,0)--(3.75,2.2);
\node[above right] at (-3.75,-2.75) {$R^{\theta_1^0}$};
\node[below right] at (-3.75,2.75) {$R^{\theta_1^1}$};
\node[above left] at (3.75,-2.75) {$R^{\theta_1^2}$};
\node[below left] at (3.75,2.75) {$R^{\theta_1^2}$};
\end{tikzpicture}
}
\caption{Left: Polygon tiling by a ReLU representation $\mathcal{M}_3\colon \R^2\to\R^2$. The first layer partitions $\mathcal \R^2$ into the $4$ unbounded polygons  $R^{\theta_1^0}, R^{\theta_1^1}, R^{\theta_1^2}, R^{\theta_1^3}$, divided by solid lines. They are further partitioned by the second layer (dashed lines). On each polygon the data is represented via a different affine linear map. Note the importance of the first layer in the architecture, determining the partitions of two polygons of the second layer up to a shift. 
Right: Polygon tiling  by a  a piecewise linear ReLU representation $\mathcal{M}_3\colon \R^2\to\R^2$, illustrating the importance of allowing affine linear maps. If, as here, all bias vectors vanish such that on each polygon the data is represented via a linear map, the domain partitioning degenerates to a tiling into cones, only one of which is being further partitioned by the second layer.
}
\label{Fig2}
\end{figure}
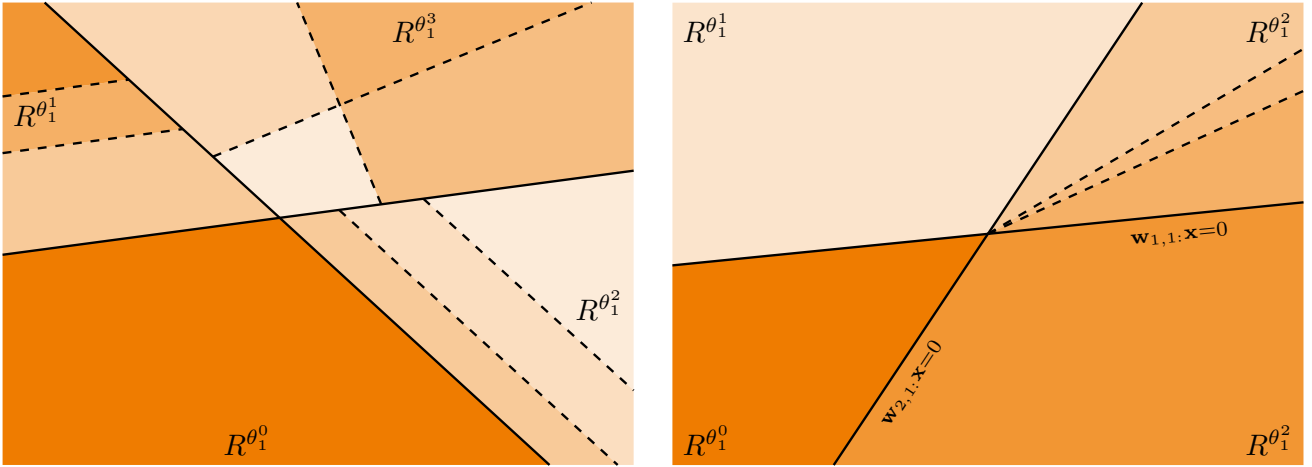

\medskip
In our terminology Proposition~\ref{lemReLU} reads as the following qualitative result, well know in the literature, e.g.~\cite{BengioNIPS2014}, and further illustrated in Figures~\ref{Fig2}.

\medskip

\begin{cor}\label{PartitionTheorem}
(i) Every ReLU representation $\mathcal{M}_L$ is a piecewise affine linear operator with respect to a partition of the input space $\mathcal X$ into convex polytopes (on each of which $\mathcal{M}_L$ is affine linear). 
The number of polytopes is equal to the number of diagonal configurations of $\mathcal{M}_L$. \\
(ii) If the biases of all layers of $\mathcal{M}_L$ vanish, then $\mathcal{M}_L$ is piecewise linear with respect to a partion of $\mathcal X$ into convex cones.
\end{cor}

\subsection{Affine linear maps}
%Deep representations partition the input space and increasing the number of layers successively refines the partition regions, each of which is associated with an affine linear transform. 
%Recall that $\theta= [\theta_1^\top, \cdots, \theta_{L-1}^\top]^\top$ ($\theta_k\in\R^{N_{k}}$) denote the configuration vector  and $\theta_k^+$ denote the set of indices of entry $1$ in $\theta_k$. 
We now give a precise characterization in terms of an atomic decomposition for the affine transform induced by a configuration. Here we refer to a rank one matrix (the outer product of two vectors) as an atom. We show that
 the atoms that linearly combine the linear part of the affine transform  induced by the configuration $\theta$ are exclusively determined by the Kronecker product of $\diag(\theta_{1})\bW_1$ and $\bW_L\diag(\theta_{L-1})$. 
Thus, increasing the number of rows of $\bW_1$ and the number of columns of $\bW_L$ (i.e., the widths of layers $1$ and $L-1$) increases the number of atoms in expressing all affine transform pieces of $\mathcal{M}_L$. 
The coefficients in the linear combination of the atoms are sums of weight products over paths between those layers. Each {path is obtained by taking} one entry from one nonvanishing column in $\bW_j\diag(\theta_{j-1})$ for $j=2,\ldots, L-1$. Increasing the widths and the number of intermediate layers, in different ways,  increases the number of paths contributing to a coefficient.

\begin{thm}\label{Thm3}
Let $\theta$ be a configuration of an $L$-layer ReLU representation $\mathcal{M}_L$. Then
the linear part of the affine linear transform $\mathcal{M}_L^{\theta}$ induced by $\theta$ is a linear combination of  atoms of the form $\{\bw_{L,: i_{L-1}}\otimes \bw_{1,i_1:}\}_{i_{L-1} \in \spt\theta_{L-1}, i_1 \in \spt\theta_1}$. 
Specifically:
\begin{itemize}
\item[(i)] For $L=2$ the linear part of $\mathcal{M}_L^{\theta}$ is the sum of  $|\spt\theta_1| $ atoms.
\item[(ii)] For $L =3$ the linear part of $\mathcal{M}_L^{\theta}$ is a linear combination of  $|\spt\theta_1| |\spt\theta_{2}|$ atoms and $w_{2,i_2i_1}$ is the coefficient for atom $\bw_{3,: i_{2}}\otimes\bw_{1,i_1:}$.
\item[(iii)] For $L>3$ the linear part of $\mathcal{M}_L^{\theta}$ the linear combination
\begin{align}\label{ReLUPaths}
\sum_{i_{L-1}\in \spt\theta_{L-1},i_1 \in \spt\theta_1}c_{i_{L-1},i_1}\bw_{L,: i_{L-1}}\otimes\bw_{1,i_1:},
\end{align}
of at most  $|\spt\theta_1|  |\spt\theta_{L-1}|$ atoms, with coefficients 
\begin{align*}
c_{i_{L-1},i_1}= 
\sum_{\substack{i_{L-2} \in \spt\theta_{L-2},\\ \ldots, i_2 \in \spt\theta_2}} w_{L-1,i_{L-1}i_{L-2}}\cdots w_{2,i_2i_1},
\end{align*}
each of which is the sum of  products consisting of at most one weight from each layer along 
$\prod_{j=2}^{L-2}|\spt\theta_{j}|$ possible paths.
\end{itemize}
%(ii) Increasing the numbers of columns of $M1$ and the rows of $M_L$ life the expression power of the induced affine linear transform.
\end{thm}
\proof (i) The linear part of $\mathcal{M}_2^{\theta}$  is  the sum of $|\spt\theta_1|$ atoms, namely
\begin{align*}
\bW_2\diag(\theta_1)\bW_1 &= \sum_{i=1}^{N_1}\bw_{2,: i}(\theta_1)_i\bw_{1,i:} 
 =  \sum_{i_1 \in \spt\theta_1}  \bw_{2,: i_1}\otimes\bw_{1,i_1:}.
\end{align*}
(ii) The $i_2$-th row of $\bW_2\diag(\theta_1)\bW_1$ is therefore $\sum_{i_1 \in \spt\theta_1} w_{2,i_2 i_1} \bw_{1,i_1:}$ and thus the linear  part of $\mathcal{M}_3^{\theta}$ is
\begin{align*}
\bW_3\diag(\theta_2)\bW_2\diag(\theta_1)\bW_1  
%= \sum_{i_2 \in \theta_2^+}\sum_{i_1 \in \theta_1^+}\bw_{3,\bullet i_2} w_{2,i_2i_1} \bw_{1,i_1\bullet}^\top
= \sum_{i_2 \in \spt\theta_2}\sum_{i_1 \in \spt\theta_1}w_{2,i_2i_1} \bw_{3,: i_2}\otimes\bw_{1,i_1:}.
\end{align*}

(iii) The $i_3$-th row of $\bW_3\diag(\theta_2)\bW_2\diag(\theta_1)\bW_1$ is
 $\sum_{i_1 \in \spt\theta_1,i_2 \in \spt\theta_2} w_{2,i_2 i_1}w_{3,i_3 i_2} \bw_{1,i_1:}$ and thus the linear  part of $\mathcal{M}_4^{\theta}$ is
\begin{align*}
\bW_4\diag(\theta_3)\bW_3\diag(\theta_2)\bW_2\diag(\theta_1)\bW_1  
= \sum_{\substack{i_3 \in \spt\theta_3, \\ i_1\in \spt\theta_1}} \left(\sum_{i_2 \in \spt\theta_2}w_{3,i_3 i_2}w_{2,i_2i_1}\right)\bw_{4,:i_3}\otimes \bw_{1,i_1:}.
\end{align*}
%Note that $\sum_{i_2 \in \theta_2^+}u_{3, i_2}(i_3)u_{2,i_1}(i_2)$ is a scalar for atom $u_{4, i_3} v_{1, i_1}^\top$, for each pair of $i_ 3 \in \theta_3^+$ and $i_1 \in \theta_1^+$. 
Successively continuing, the linear part of $\mathcal{M}_L^{\theta}$ is  \eqref{ReLUPaths}.
\qed

\medskip

On the polytope $R^{\theta}$ we therefore obtain the following expression for $\mathcal{M}_L$:
\begin{align}
\mathcal M_L^{\theta}\bx 
%& = M_L\bD_{L-1}^{\theta}\cdots \bD_2^{\theta}M_2(\bD_1^{\theta}\bW_1\bx + \bD_1^{\theta}\bb_1) \\
%& = M_L\bD_{L-1}^{\theta}\cdots \bD_3^{\theta}M_3(\bD_2^{\theta}\bW_2\bD_1^{\theta}\bW_1\bx + \bD_2^{\theta}\bW_2\bD_1^{\theta}\bb_1 +\bD_2^{\theta}\bb_2) \\
%& = M_LD_{L-1}^{\theta}\cdots  D_4^{\theta}M_4  (D_3^{\theta}\bW_3D_2^{\theta}\bW_2D_1^{\theta}\bW_1\bx + D_3^{\theta}\bW_3D_2^{\theta}\bW_2\bD_1^{\theta}b_1 +D_3^{\theta}\bW_3D_2^{\theta}\bb_2) \\
& = \bW_L\diag(\theta_{L-1})\cdots \diag(\theta_1)\bW_1\bx + \bb \label{affinepiece}\\
%& = \mathcal L(M_L\bD_{L-1}^{\theta}\cdots\bD_2^{\theta}M_1)\bx+ \sum_{j=1}^{L-1}\bW_{L}D_{L-1}\cdots \bW_{j+1}D_j^{\theta}b_j + \bb_L.
&=  \sum_{i_{L-1} \in \spt\theta_{L-1}} \alpha_{L, i_{L-1}}(\bx) \bw_{L,: i_{L-1}} + \bb,   \nonumber
\end{align}
where 
\begin{align*}
\bb = \sum_{j=1}^{L-1}\bW_{L}\diag(\theta_{L-1})\cdots \bW_{j+1}\diag(\theta_j)\bb_j + \bb_L,
\end{align*}
and where
the coefficients $\alpha_{L, i_{L-1}}(\bx)$ associated with the column $\bw_{L,: i_{L-1}}$ of $\bW_L\diag(\theta_{L-1})$ are
%\begin{align*} %\label{ReLUcoeff}
%\alpha_{L, i_{L-1}}(\bx) = 
%\begin{cases}
%\sum\limits_{i_1 \in \spt\theta_1} \bw_{1,i_1\bullet}^\top \bx, &L = 2 \\
%\sum\limits_{i_1 \in \spt\theta_1}w_{2,i_2i_1}\bw_{1,i_1\bullet}^\top\bx, &L=3 \\
%\sum\limits_{i_{L-2}\in \spt\theta_{L-2},\ldots,i_1 \in \spt\theta_1} w_{L-1,i_{L-1}i_{L-2}}\cdots w_{2,i_2i_1}\bw_{1,i_1\bullet}^\top \bx, &L>3,
%\end{cases}
%\end{align*}
\begin{align*} 
\alpha_{2, i_{1}}(\bx)=\bw_{1,i_1:} \bx
\end{align*}
in the case of $L=2$ layers;
\begin{align*} %\label{ReLUcoeff}
\alpha_{3, i_{2}}(\bx)&=\sum\limits_{i_1 \in \spt\theta_1}w_{2,i_2i_1}\bw_{1,i_1:}\bx  
= \bw_{2,i_2:}\diag(\theta_1)\bW_1\bx
\end{align*}
in the case of $L=3$ layers;
%
%\begin{align*} %\label{ReLUcoeff}
%&\sum\limits_{i_1 \in \spt\theta_1} \bw_{1,i_1:}^\top \bx  &\text{ for }L = 2, \\
%&\sum\limits_{i_1 \in \spt\theta_1}w_{2,i_2i_1}\bw_{1,i_1:}^\top\bx  &  \text{ for }L=3,
%\end{align*}
and 
\begin{align*}
\alpha_{L, i_{L-1}}(\bx)
&=\sum_{\substack{i_{L-2}\in \spt\theta_{L-2},\\ \ldots,i_1 \in \spt\theta_1}} w_{L-1,i_{L-1}i_{L-2}}\cdots w_{2,i_2i_1}\bw_{1,i_1:} \bx\\
&=\bw_{L-1,i_{L-1}:}\diag(\theta_{L-2})\bW_{L-2}\cdots\diag(\theta_1)\bW_1\bx
\end{align*}
in the general case of $L>3$ layers. In particular, the affine transform $\mathcal M_L^{\theta}$ maps its domain $R^{\theta}$ into
\[ 
\linhull\{\bw_{L,:i_{L-1}}\colon{i_{L-1} \in \spt\theta_{L-1}}\}+\bb.
\]

As an immediate application we estimate a Lipschitz bound for $\mathcal{M}_L^{\theta}$, which can be interpreted as a measure for the gain of local input perturbations to that of the outputs of $\mathcal{M}_L$ on $R^{\theta}$. The bound depends on the number of activated rectifiers.
Given the atomic representation, it may have benefits to normalize the columns of $\bW_L$ and the rows of $\bW_1$, depending, e.g., on whether the model is used for signal analysis or synthesis. The following result can easily be modified for the case without this normalization assumption.

\begin{thm}\label{ddformula}
Let $\theta$ be a configuration of an $L$-layer ReLU representation $ \mathcal M_L$ with $L> 3$, and let $\bx_1,\bx_2\in R^{\theta}$.
\begin{itemize}
\item[(i)]
Suppose that $\bW_L$ has normalized columns, that $\bW_1$ has normalized rows, and let $C$ be the maximum of  the absolute value of all weights in $\bW_2,\ldots,\bW_{L-1}$. 
 Then
\begin{align*}
\| \mathcal M_L^{\theta}(\bx_1) - \mathcal M_L^{\theta}(\bx_2) \|_{2} \le  
\left(C^{L-2} \prod_{k=1}^{L-1}|\spt\theta_{k}|\right) \|\bx_1 - \bx_2\|_{2}. 
\end{align*}
If $N = \max\{N_1,\ldots,N_{L-1}\}$ then   $(CN)^{L-2} N$ is a global Lipschitz bound for $\mathcal{M}_L$.
\item[(ii)] If $\sigma$ is the maximum of the spectral norms of the weight matrices, then 
\[
\| \mathcal M_L^{\theta}(\bx_1) - \mathcal M_L^{\theta}(\bx_2) \|_{2} \le  
\sigma^L \|\bx_1 - \bx_2\|_{2}. 
\]
\end{itemize}
\end{thm}
\proof Under the assumptions of (i), for $\bx_1, \bx_2\in R^{\theta}$ we get 
\begin{align*}
\|\mathcal M_L^{\theta}(\bx_1) - \mathcal M_L^{\theta}(\bx_2)\|_{2}  
& \le \sum_{i_{L-1}\in \spt\theta_{L-1}} |\alpha_{L, i_{L-1}}(\bx_1-\bx_2)| \\
& \le \sum_{\substack{i_{L-1}\in \spt\theta_{L-1},\\\ldots,i_1 \in \spt\theta_1}}C^{L-2}|\bw_{1,i_1:} (\bx_1 - \bx_2)| \\
& \le \sum_{\substack{i_{L-1}\in \spt\theta_{L-1},\\\ldots,i_1 \in \spt\theta_1}} C^{L-2} \|\bx_1 - \bx_2\|_{2} \\
& \le \left( C^{L-2} \prod_{k=1}^{L-1}|\spt\theta_{k}| \right)\|\bx_1 - \bx_2\|_{2}.
\end{align*}
For the global Lipschitz estimate note that,
since $|\spt\theta_k|\leq N_k$, we have
\[
C^{L-2}\prod_{k=1}^{L-1}|\spt\theta_k| \le C^{L-2} N^{L-1}.
\]
Part (ii) follows directly from~\eqref{affinepiece}.
\qedhere
%\medskip
%If $C > 1$, the Lipschitz parameter tends to infinity as the number of layers increases.
%Recall that increasing the number of layers refines the partitioning of the input space and thus reduces the size of the domains of the affine linear transforms. To achieve stability with respect to local perturbations we must therefore have $C<1$ and a sufficiently large number of layers $L$. Indeed, if $C < 1$, then the Lipschitz parameter is decreasing as the number of layers increases and $L > \frac{\ln |\theta_1^+|}{p \ln \frac{1}{C}} 
%+ 2 $, where $p := \min_{i=2,\ldots,L-1}|\theta_i^+| \neq 0$, guarantees $|\theta_1^+| C^{\sum_{i=2}^{L-1}|\theta_{i}^+|} < 1$. This result is consistent with the stability of the scattering transform to local deformations, implied from the fact that the Lipschitz continuity locally linearizes local deformations at large $L$ in the scattering transform \cite{..}.
%
%

\medskip
It is clear that the global Lipschitz bound $(CN)^{L-2}N$ derived for $\mathcal{M}_L$ via the above crude estimate from its affine linear pieces  is far from being optimal. 
%A similar global Lipschitz bound has also been derived in \cite{CSGenMod2017}. 
As such, Theorem~\ref{ddformula} can be regarded as a refinement of a similar global Lipschitz bound derived in~\cite{CSGenMod2017}. 
The fact that increasing the number $L$ of layers of the representation refines the partitioning of the input space, implies that, in order to keep stability, the Lipschitz bound for $\mathcal M_L$ should be a non-increasing function of $L$; otherwise a tiny part of the input space could cause instability of the representation. We are thus particularly interested in deriving a sufficient condition for the Lipschitz bound to not be an increasing function of $L$. Achieving this for the bound in (i) requires
%If $C \leq P^{\frac{1}{L-2}}/N$, then
%\[
%C^{L-2}N^{L-1} \le P N.
%\]
%To achieve a small Lipschitz parameter for $\mathcal M_L$ requires that $P <1$, in which case $P^{\frac{1}{L-2}}$ is an increasing function of $L$, with limit $1$ as $L\to\infty$. Thus  
%{\bf (in particular? already clear from theorem)}, 
$C \leq 1/N$, i.e., to achieve a stable representation regardless the number of layers requires the mean and variance of the weight coefficients to be very small at large $N$.
This might be related to the batch normalization technique in learning
DNNs~\cite{Ioffe2015BatchNA%introduction and heuristic attribution to reduction of covariance shift
,Santurkar2018HowDB%Alterntive intuition as smoothing gradient of loss function in
}. On the other hand, the suffiecient condition of having the spectral norms of the weight matrices not exceed $1$ can be achieved via optimization techniques by imposing the Frobenius  norms of the weight matrices to not exceed $1$.

With regards to Theorem~\ref{ddformula}(i),
we remark on two observations further suggesting that asymptotic stability of the  Lipschitz bound for $\mathcal{M}_L$ for large number of layers plays a role in the learning process of function approximation via deep feedforward neural networks.
The back-propagation algorithm, designed to carry out the learning task, is based on (sub)gradient descent in the landscape of a loss function $\mathcal{L}$ in the network parameter space. It is believed that in the course of training, both, the maximum magnitude component and the smoothness of the gradient of $\mathcal L$ affect the learning performance. \\
We first consider the maximum magnitude component of the gradient.
Let $\mathcal L^{\theta}$ denote the restriction of the loss function to $R^{\theta}$ and for $\bx \in R^{\theta}$ denote
$\by_k= \mathcal M_k^{\theta}(\bx)$ and $\ba_k = \diag(\theta_k)\by_k$. Then
\begin{align} \label{range}
\frac{\partial \mathcal L^{\theta}}{\partial \by_k} = \Sigma^{\theta_k}(\by_{k})\bW_{k+1}^\top\Sigma^{\theta_{k+1}}(\by_{k+1}) \cdots \bW_{L}^\top \nabla_{\by_L}\mathcal L^{\theta},
\end{align}
where $\Sigma^{\theta_l}(\by_{l})$, for $l=k, \ldots, L-1$, is a diagonal matrix with entries $0$ or $1$, corresponding to the value of the directional derivative of the rectifier function, which is $\partial{a_{l,i}}/\partial y_{l,i}=1$ if $y_{l,i}  \ge 0$ and $0$ otherwise.
%\begin{align*}
%\frac{\partial{a_{l,i}}}{\partial y_{l,i}} = 
%\begin{cases}
%1 &\text{ if $y_{l,i}  \ge 0$,} \\
%0 &\text{ else.}
%\end{cases}
%\end{align*}
Note that this value is $1$ at $y_{l,i} = 0$, since the subdifferential of the one-dimensional rectifier at $y_{l,i} = 0$ is the interval $[0, 1]$ and the directional derivative of a one-dimensional  convex function is the maximum of the subdifferential. 
%With  $\|B\|_{\max} := \max_{i,j} |b_{i,j}|$ for $B\in\R^{m\times n}$, and recalling 
Using the estimate $\|B\| \leq \sqrt{mn} \|B\|_{\max}$ for $B\in\R^{m\times n}$ on the weight matrices,  the maximum magnitude entry  of $\partial\mathcal L^{\theta}/\partial \by_k$ is bounded by
\begin{align}\label{argument1}
\left \|\frac{\partial \mathcal L^{\theta}}{\partial \by_k}\right\|
& \le \|\bW_{k+1}^\top \| \cdots  \|\bW_{L}^\top\| \|\nabla_{\by_L}\mathcal L^{\theta}\|  \nonumber\\
& \le (NC)^{L-k} \|\nabla_{\by_L}\mathcal L^{\theta}\|.
\end{align}
%The $(NC)^{L-k}$ pertains to the upper bound on the range of gradient at layer $k$. 
If $NC > 1$, then  $(NC)^{L-k}$ increases when $k$ decreases. This implies that the maximum magnitude entries of the gradient at early layers can potentially have larger variations, which would hamper the learning performance. \\
Our second consideration concerns the smoothness of the gradient of the loss function. Globally this gradient is notoriously nonsmooth and thus again we restrict to the individual polytope regions $R^{\theta}$. Assume that  there the gradient of the loss function is $\beta_{\theta}$-smooth, i.e., suppose that for all $\by_k= \mathcal M_k^{\theta}(\bx)$ and 
$\by'_k= \mathcal M_k^{\theta}(\bx')$, where $\bx,\bx' \in R^{\theta}$, the estimate
\[ \|\nabla_{\by} \mathcal L^{\theta} - \nabla_{\by'} \mathcal L^{\theta}\| \le \beta_{\theta} \| \by - \by' \|\]
holds.
For any layer $k$, \eqref{range} implies
\begin{align}\label{argument2}
\left\|\frac{\partial \mathcal L^{\theta}}{\partial \by_k} - \frac{\partial \mathcal L^{\theta}}{\partial \by'_k}\right\| & \leq (NC)^{L-k}  \|\nabla_{\by_L}\mathcal L^{\theta} -  \nabla_{\by'_L}\mathcal L^{\theta}\|\\ \nonumber 
& \leq \beta_{\theta}  (NC)^{L-k}\|\by_L - \by'_L\|.
\end{align}
Similar to \eqref{argument1}, here a condition like $C\leq 1/N$ is needed to guarantee to avoid blowing up of the Lipschitz parameters at early layers during learning. \\
%Controlling the maximum magnitude of all weights is of course but one, rather crude, among many  empirically successful ideas to achieve stability in learning. Following the above derivations one may also attempt to control, say, the Frobenius norms of the weight matrices. 
We hope that having precise expressions such as \eqref{ReLUPaths} for deep representations can contribute to paving a way to develop new and to better understand existing regularization techniques such as batch normalization, dropout \cite{HintonDropout1,HintonDropout2,GoodfellowDropout} or deep residual learning \cite{He16}.

We finally would like to make one more signal processing related remark. The smoothness of a loss function not only relates to the stability in learning a deep representation, it also relates to deriving local minimizers of $\mathcal{L}$ over the input space $\mathcal{X}$ using gradient descent. Following \eqref{argument2}, we have
\begin{align*}
\left\|\frac{\partial \mathcal L^{\theta}}{\partial \by_k} - \frac{\partial \mathcal L^{\theta}}{\partial \by'_k}\right\| & \leq   \beta_{\theta}  (NC)^{L-k} \|\mathcal{M}_L^{\theta}(\bx- \bx')\|\\
& \leq \beta_{\theta} (NC)^{L-k} \|\bW_{L}\|  \cdots \|\bW_1\|\|\bx- \bx'\| \\
& \leq \beta_{\theta} (NC)^{2L-k} \|\bx- \bx'\|.
\end{align*}
%If $\mathcal L$ retains global $\beta$-smoothness, then for any layer $k$, the gradient of $\mathcal L$ on the input $\by_k(\bx)$ to the $k$-th hidden layer inherits $\beta (NC)^{2L-k}$-smoothness. By chain rule,
Since
\[\frac{\partial \mathcal L^{\theta}}{\partial \bx} = \bW_1^\top \frac{\partial \mathcal L^{\theta}}{\partial \by_1},
\] 
this implies
\begin{align*}
\left\|\frac{\partial \mathcal L^{\theta}}{\partial \bx} - \frac{\partial \mathcal L^{\theta}}{\partial \bx'}\right\| \leq \beta_{\theta} (NC)^{2L} \|\bx - \bx'\|,
\end{align*}
i.e., here again $C \leq 1/N$ is sufficient to stabilize the smoothness of the gradient for large $L$.

\bibliography{DeepRepresentationReLUNN}
\bibliographystyle{ieeetr}
\end{document}